\documentclass[runningheads]{llncs}
\usepackage{graphicx}

\usepackage{tikz}
\usepackage{comment}
\usepackage{amsmath,amssymb} 
\usepackage{color}
\usepackage[colorlinks,linkcolor=blue]{hyperref}

\usepackage{cite}
\usepackage{subfigure}
\usepackage{multirow}
\usepackage{diagbox}
\usepackage{rotating}
\usepackage{booktabs}
\usepackage{colortbl}
\usepackage{overpic}
\usepackage{textcomp}
\usepackage{contour}
\usepackage{algorithm}
\usepackage{algorithmic}

\usepackage[accsupp]{axessibility}  

\newcommand{\figref}[1]{Fig. \ref{#1}}
\newcommand{\tabref}[1]{Table. \ref{#1}}

\newcommand{\secref}[1]{Sec. \ref{#1}}

\newcommand{\ourmodel}{OSFormer}

\renewcommand{\tabcolsep}{.5mm}

\newcommand{\eg}{\emph{e.g.}}
\newcommand{\ie}{\emph{i.e.}}
\newcommand{\etal}{\emph{et al.}}

\graphicspath{{./Imgs/}}

\begin{document}
\pagestyle{headings}
\mainmatter
\def\ECCVSubNumber{1317}  

\title{OSFormer: One-Stage Camouflaged Instance Segmentation with Transformers} %

\titlerunning{OSFormer}
%
\author{Jialun~Pei$^\dagger$\inst{1}\orcidID{0000-0002-2630-2838} \and
Tianyang~Cheng$^\dagger$\inst{2}\orcidID{0000-0001-7910-3624} \and
Deng-Ping~Fan$^*$\inst{3}\orcidID{0000-0002-5245-7518} \and
He Tang\inst{2}\orcidID{0000-0002-8454-1407} \and 
Chuanbo~Chen\inst{2}\orcidID{0000-0001-8006-7851} \and
Luc~Van Gool\inst{3}\orcidID{0000-0002-3445-5711}\index{Van Gool, Luc.}}
\authorrunning{Pei et al.}
%
\institute{School of Computer Science and Technology, HUST, China \and
School of Software Engineering, HUST, China \and
Computer Vision Lab, ETH Zurich, Switzerland 
}
\maketitle
\newcommand\blfootnote[1]{%
\begingroup
\renewcommand\thefootnote{}\footnote{#1}%
\addtocounter{footnote}{-1}%
\endgroup
}

\begin{abstract}
We present \textbf{OSFormer}\blfootnote{$\dagger$ Equal contributions. 
* Corresponding author (\email{dengpfan@gmail.com}).}, the first one-stage transformer framework for camouflaged instance segmentation (CIS). \ourmodel~is based on two key designs. 
First, we design a \textbf{location-sensing transformer} (LST) to obtain 
the location label and instance-aware parameters 
by introducing the location-guided queries and the blend-convolution feed-forward network.
Second, we develop a \textbf{coarse-to-fine fusion} (CFF) to merge diverse context information from the LST encoder and CNN backbone. 
Coupling these two components enables OSFormer to efficiently blend local features and long-range context dependencies for predicting camouflaged instances. 
Compared with two-stage frameworks, our OSFormer 
reaches 41\% AP and achieves good convergence efficiency without requiring enormous training data, \ie, only 3,040 samples under 60 epochs. 
Code link: \url{https://github.com/PJLallen/OSFormer}.

\keywords{Camouflage, instance segmentation, transformer}
\end{abstract}

\section{Introduction}
Camouflage is a powerful and widespread means of avoiding detection or recognition that stems from biology \cite{troscianko2021variable}. In nature, camouflage objects have evolved a suite of concealment strategies to deceive perceptual and cognitive mechanisms of prey or predators, such as background matching, self-shadow concealment, obliterative shading, disruptive coloration, and distractive markings \cite{cuthill2019camouflage,stevens2009animal}. 
These defensive behaviors make camouflaged object detection (COD) a very challenging task compared to generic object detection~\cite{ren2015faster,redmon2016you,lin2017focal,tian2019fcos,carion2020end}. COD is dedicated to distinguishing camouflaged objects with a high degree of intrinsic similarity with backgrounds~\cite{fan2020camouflaged}. It is essential to use computer vision models to assist human visual and perceptual systems for COD, such as polyp segmentation~\cite{fan2020pranet,ji2021progressively}, lung infection segmentation~\cite{fan2020inf}, wildlife protection, and recreational art~\cite{chu2010camouflage}.

Thanks to the build of large-scale and standard benchmarks like COD10K~\cite{fan2020camouflaged}, CAMO~\cite{le2019anabranch}, CAMO++~\cite{ltnghia-TIP2022}, and NC4K~\cite{yunqiu_cod21}, the performance of COD has received significant progress. 
However, COD only separates camouflaged objects from the scene at region-level while ignoring further instance-level identification. 
Recently, Le \etal~\cite{ltnghia-TIP2022} presented a new camouflaged instance segmentation (CIS) benchmark and a camouflage fusion learning framework.
Capturing camouflaged instances can provide more clues (\eg, semantic category, the number of objects) in real-world scenarios, thus CIS is more challenging. 

\begin{figure}[t!]
\centering
\includegraphics[width=.98\linewidth]{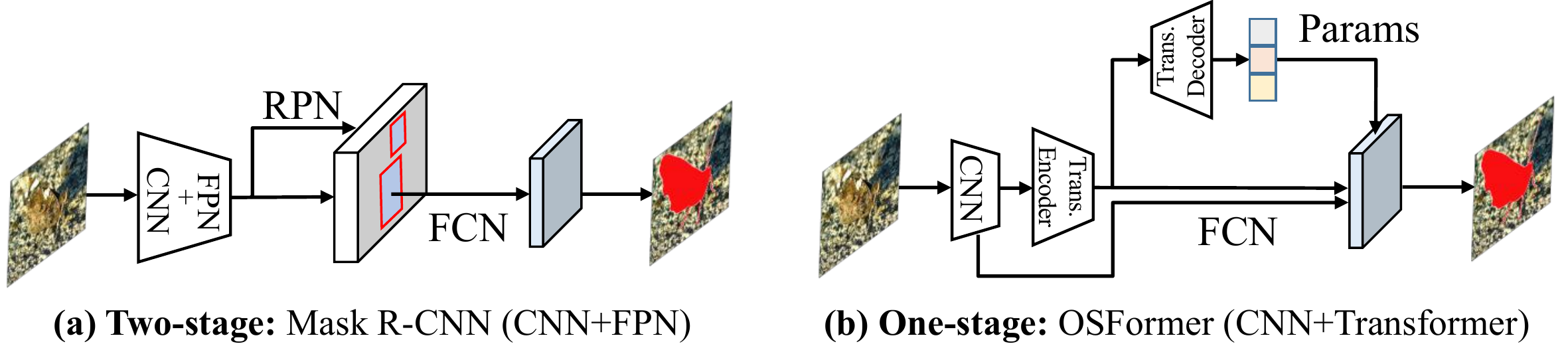}
\caption{Framework comparisons of Mask R-CNN~\cite{he2017mask} and the proposed~\ourmodel.}\label{fig:Cmp}
\end{figure}

Compared to generic instance segmentation~\cite{he2017mask}, CIS needs to be 
performed in more complex scenarios with high feature similarity and results in class-agnostic masks. Moreover, various instances may display different camouflage strategies in a scene, and they may combine to form mutual camouflage. 
These derived ensemble camouflages make the CIS task even more daunting. 
When humans gaze at a heavily camouflaged scene, the visual system will instinctively sweep across a series of local scopes throughout the whole scene to search for valuable clues~\cite{sandon1990simulating,matthews2015human}. Inspired by this visual mechanism, we present a novel location-aware CIS approach that meticulously captures crucial information at all positions (\ie, \textit{local context}) in a \textit{global perspective} and directly generates camouflaged instance masks (\ie, \textit{one-stage} model). 

Thanks to the rise of the transformer~\cite{vaswani2017attention} in the visual domain, we can employ self-attention and cross-attention to capture long-range dependencies and build global content-aware interactions~\cite{carion2020end}. Although the transformer model has shown strong performance on some dense prediction tasks \cite{wang2021pyramid,xie2021segformer,guo2021sotr,wang2021max}, it requires embracing large-scale training data and longer training epochs. 
However, there is currently only limited instance-level training data available as a brand-new downstream task. To this end, we propose a \textbf{location-sensing transformer (LST)} based on~\cite{zhu2020deformable} to achieve faster convergence and higher performance with fewer training samples. To dynamically yield location-guided queries for each input image, we grid the multi-scale global features output from the LST encoder into a set of feature patches with varying local information. Compared to 
zero initialization of object queries in vanilla DETR~\cite{carion2020end}, the proposed location-guided queries can lead to focus on location-specific features and interact with global features through cross-attention to gain the instance-aware embeddings.
This design effectively speeds up convergence and significantly improves detecting camouflaged instances. To enhance local perception and the correlation between neighboring tokens, we introduce convolution operations into the standard feed-forward network \cite{vaswani2017attention}, which we term blend-convolution feed-forward network (BC-FFN). Therefore, our LST-based model can seamlessly integrate local and global context information and efficiently provide location-sensitive features for segmenting camouflaged instances.

In addition, we design a \textbf{coarse-to-fine fusion (CFF)} to integrate multi-scale low- and high-level features successively derived from ResNet~\cite{he2016deep} and LST to bring out the shared mask feature. Since the edges of the camouflaged instances are difficult to capture, a reverse edge attention (REA) module is embedded in our CFF module to enhance the sensitivity to edge features. 
Finally, inspired by~\cite{huang2017arbitrary}, we introduce the dynamic camouflaged instance normalization (DCIN) to generate the masks by uniting the high-resolution mask feature and the instance-aware embeddings.
Based on those mentioned above two novel designs, \ie, LST and CFF, we provide a new one-stage framework \textbf{\ourmodel} for camouflaged instance segmentation (\figref{fig:Cmp}).
To the best of our knowledge, \ourmodel~is the first work to explore the transformer-based framework for 
the CIS task. Our \textbf{contributions} are as follows:
\begin{enumerate}
\item We propose \textbf{\ourmodel}, the first one-stage transformer-based framework designed for the camouflage instance segmentation task. It is a flexible framework that can be trained in an end-to-end manner. 

\item We present a \textbf{location-sensing transformer (LST)} to dynamically seize instance clues at different locations. Our LST contains an encoder with the blend-convolution feed-forward network to extract multi-scale global features and a decoder with the proposed location-guided queries to bring the instance-aware embeddings.
The proposed LST structure converges quickly with limited \ie, about 3,000 images, training data.

\item A novel \textbf{coarse-to-fine fusion (CFF)} is proposed to get the high-resolution mask features by fusing multi-scale low- and high-level features from the backbone and LST block. In this module, reverse edge attention (REA) is embedded to highlight the edge information of camouflaged instances.

\item Extensive experiments show that \ourmodel~performs well for the challenging CIS task, \textbf{outperforming} 11 popular instance segmentation approaches by a large margin, \eg, \textit{8.5\% AP improvement} on the COD10K test set.

\end{enumerate}

\section{Related Work}

\noindent\textbf{Camouflaged Object Detection.} 
This category of the models aims to identify objects that blend in the surrounding scene~\cite{fennell2021camouflage}. Earlier studies mainly employed low-level handcrafted contrast features and certain heuristic priors (\eg, color \cite{huerta2007improving}, texture \cite{bhajantri2006camouflage,song2010new} and motion boundary \cite{mondal2020camouflaged}) to build camouflaged object detection (COD) models. With the popularity of deep learning architecture and the release of large-scale pixel-level COD datasets~\cite{le2019anabranch,fan2020camouflaged}, the performance of COD has been improved by leaps and bounds in the past two years. 
Deep learning methods \cite{mei2021camouflaged,zhu2021inferring,yan2021mirrornet,ren2021deep} utilize CNNs to extract high-level informative features to search and locate camouflaged objects, and then design an FCN-based decoder to purify features to predict camouflaged maps. For instance, Mei \etal \cite{mei2021camouflaged} presented a positioning and focus network (PFNet) to mimic the process of predation in nature. PFNet first leverages the positioning module to locate the potential targets and use the focus module to refine the ambiguous regions. Zhai \etal \cite{zhai2021mutual} adopted a mutual graph learning strategy to train the regions and edges of camouflaged objects interactively. Afterward, Lyu \etal \cite{yunqiu_cod21} proposed a ranking network that simultaneously localizes, segments, and ranks concealed objects for better prediction. Recently, a novel uncertainty-guided transformer-based model proposed by Yang \etal \cite{yang2021uncertainty} is designed to infer uncertain regions with Bayesian learning. The COD task ignores the instance-level predicted maps essential for actual application scenarios despite the rapid development. Thus, we are dedicated to advancing the COD task from region-level to instance-level.

\noindent\textbf{Generic Instance Segmentation.}
Existing works can be roughly summarized as the top-down and bottom-up patterns. The former model performs a classic detect-then-segment design that first detects ROIs by bounding boxes and then segment pixel-level instances locally~\cite{tian2020conditional}. 
The typical model is Mask R-CNN~\cite{he2017mask}, which extends Faster R-CNN~\cite{ren2015faster} by adding a mask branch to predict instance-level masks. 
On this basis, Mask Scoring R-CNN~\cite{huang2019mask} introduces a MaskIoU head to assess the quality of the instance mask. To enhance the feature pyramid and shorten the information flow, PANet~\cite{liu2018path} creates a bottom-up path augmentation. Furthermore, Chen \etal~\cite{chen2019hybrid} proposed Hybrid Task Cascade (HTC) to interweave detection and segmentation features for joint processing. Different from the above-mentioned two-stage models, YOLACT~\cite{bolya2019yolact} is a real-time one-stage framework that embraces two parallel tasks: producing non-local prototype masks and predicting a set of mask coefficients. 

In contrast to the top-down manner, the bottom-up methods first learn instance-aware holistic embeddings and then identify each specific instance with clustering operations \cite{chen2017deeplab,liu2017sgn}. Bai \etal \cite{bai2017deep} proposed an end-to-end boundary-aware deep model derived from the classical watershed transform. SSAP \cite{gao2019ssap} can jointly learn the pixel-level semantic class and instance differentiating by an instance-aware pixel-pair affinity pyramid. However, the performance of previous bottom-up models is inferior to top-down models because of the suboptimal pixel grouping. To this end, Tian \etal \cite{tian2020conditional} presented a dynamic instance-aware network that directly outputs instance masks in a fully convolutional paradigm. The simpler strategy is efficient and performs favorably against Mask R-CNN-like frameworks. Furthermore, SOLO \cite{wang2020solo,wang2020solov2} detects the center location of instances by semantic categories and decouples the mask prediction into the dynamic kernel feature learning. Inspired by this strategy, we design a location-aware network based on the transformer to dynamically perceive camouflaged instances.

\noindent\textbf{Vision Transformer.}
Transformer \cite{vaswani2017attention} was born out of natural language processing and has been successfully extended to the field of computer vision \cite{dosovitskiy2020image}. 
The core idea of the transformer encoder-decoder architecture is a self-attention mechanism that builds long-range dependencies and captures global context information from an input sequence. 
Recently, Carion \etal~proposed DETR~\cite{carion2020end}, which combined transformer with CNN backbone to 
aggregate object-related information and provided a group of object queries to output the final set of predictions. Despite DETR pioneering a novel and concise paradigm, it still suffers from the high computational cost and the slow convergence. Considering these issues, many efforts focused on how to develop a more efficient DETR architecture~\cite{zhu2020deformable,dai2021dynamic,dai2021up}. 
Zhu \etal~\cite{zhu2020deformable} introduced a deformable attention layer embedded in the self-attention module to reduce the computational cost and training schedule. UP-DETR~\cite{dai2021up} leveraged a novel unsupervised pretext task to pre-train the transformer of DETR for accelerating convergence. However, most existing transformer models are adapted to vision tasks with many training data. 
Therefore, for downstream tasks with only small datasets, fully utilizing the performance of transformer is an urgent issue to be solved. 
To this end, we present an efficient location-sensing transformer (LST) based on the deformable DETR~\cite{zhu2020deformable} for CIS.
The proposed transformer converges easily on the CIS task with only 3,040 training samples.

\begin{figure}[t!]
\centering
\includegraphics[width=.98\linewidth]{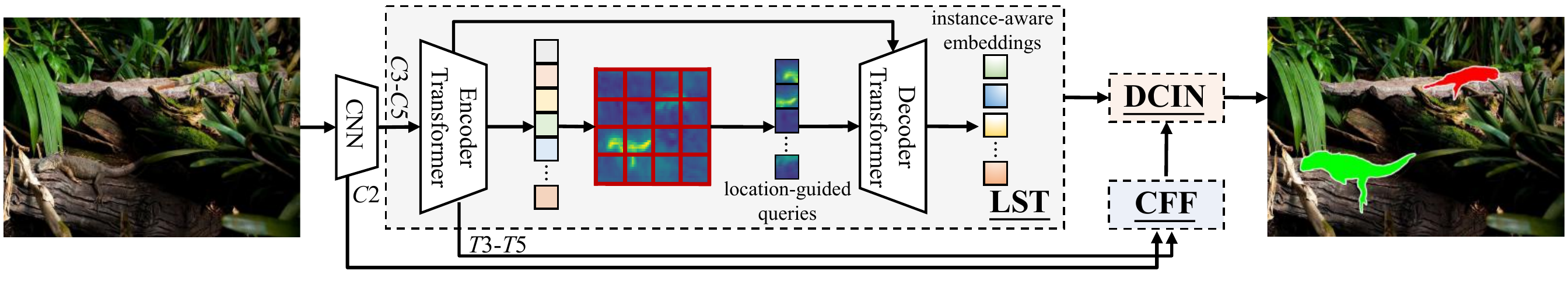}
\caption{\ourmodel~includes the location-sensing transformer (LST), coarse-to-fine fusion (CFF) module, and dynamic camouflaged instance normalization (DCIN) module. 
}
\label{fig:OSFormer}
\end{figure}

\section{\ourmodel}
\textbf{Architecture.}
The proposed \ourmodel~comprises four essential components: 
(1) a CNN backbone to extract object feature representation, 
(2) a location-sensing transformer (LST) that utilizes the global feature and location-guided queries to produce the instance-aware embeddings.
(3) a coarse-to-fine fusion (CFF) to integrate multi-scale low- and high-level features and yield a high-resolution mask feature, and 
(4) a dynamic camouflaged instance normalization (DCIN) that is applied to predict the final instance masks. We illustrate the whole architecture in \figref{fig:OSFormer}.

\subsection{CNN Backbone}
Given an input image $I\in\mathbb{R}^{H\times W\times 3}$, we use multi-scale features \{$Ci$\}$_{i=2}^{5}$ from the CNN backbone (\ie, ResNet-50~\cite{he2016deep}). 
To reduce the computational cost, 
we directly flatten and concatenate the last three feature maps ($C3, C4, C5$) into a sequence $X_{m}$ with 256 channels as input to the proposed LST encoder (\secref{sec:LST_encoder}). For $C2$ feature, we feed it into our CFF (\secref{sec:CFF}) module as high-resolution low-level feature to capture more camouflaged instance cues. 

\begin{figure}[t!]
\centering
\includegraphics[width=.98\linewidth]{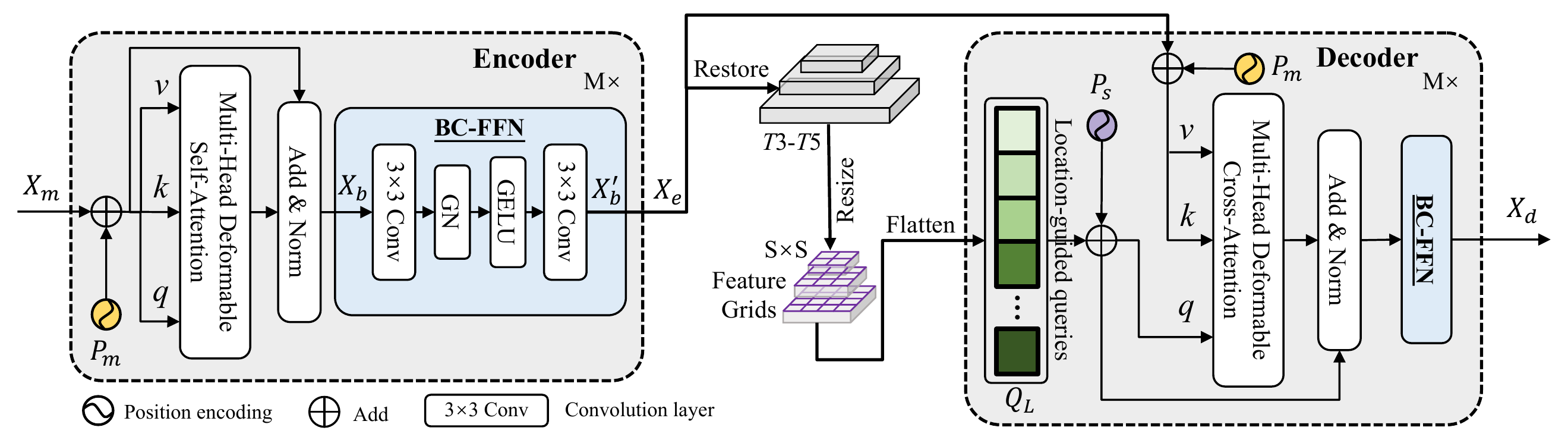}
\caption{Structure of our location-sensing transformer.}
\label{fig:LST}
\end{figure}

\subsection{Location-Sensing Transformer}\label{sec:LST_encoder}
Although the transformer can better extract global information by the self-attention layer, it requires large-scale training samples and high computational cost to support. Due to the limited data size of CIS, our goal is to design an efficient architecture that can converge faster and achieve competitive performance. In \figref{fig:LST}, we present our location-sensing transformer (LST).

\noindent\textbf{LST Encoder.} 
Unlike DETR~\cite{carion2020end} with only a single-scale low-resolution feature input to the transformer encoder, our LST encoder receive 
multi-scale features $X_{m}$ to obtain rich information.
Following deformable self-attention layers~\cite{zhu2020deformable} to better capture local information and enhance the correlation between neighboring tokens, we bring the convolution operations into the feed-forward network, named blend-convolution feed-forward network (BC-FFN). First, the feature vector is restored to the spatial dimension depending on the shape of $Ci$. Then, a convolution layer with the kernel size of $3\times3$ is performed to learn the inductive biases. Finally, we add a group normalization $(GN)$ and a $GELU$ activation to form our feed-forward network. After a $3\times3$ convolution layer, we flatten the features back into a sequence. 
%
Compared to mix-FFN~\cite{xie2021segformer}, our BC-FFN contains no MLP operations and residual connections. Unlike~\cite{wu2021cvt} that designs a convolutional token embedding at the beginning of each stage and employs depth-wise separable convolution operation in the transformer block, we only bring two convolution layers in BC-FFN.
Specifically, given an input feature $X_{b}$, the process of $\textit{BC-FFN}$ can be formulated as: 
\begin{equation}
X_{b}^{\prime}=Conv^{3}(GELU(GN(Conv^{3}(X_{b})))),
\end{equation}
where $Conv^{3}$ is a $3\times3$ convolution operation. Overall, a LST encoder layer is described as follows:
\begin{equation}
X_{e}=\textit{BC-FFN}(LN((X_{m}+P_{m})+\textit{MDAttn}(X_{m}+P_{m}))),
\end{equation}
where $P_{m}$ is denoted as the position encodings. \textit{MDAttn} and $LN$ represent multi-head deformable self-attention and layer normalization, respectively.

\noindent\textbf{Location-Guided Queries.} 
Object queries play a critical role in the transformer architecture \cite{carion2020end}, which are used as the initial input to the decoder and attain the output embeddings through the decoder layers. However, one of the reasons for the slow convergence of the vanilla DETR is that object queries are zero-initialized. To this end, we propose location-guided queries that take advantage of multi-scale feature maps $Ti, i=3,4,5$ from the LST encoder\footnote{We split and restore the $X_{e}$ to the 2D representations $T3\in\mathbb{R}^{{\frac{H}{8}}\times{\frac{W}{8}}\times D}$, $T4\in\mathbb{R}^{{\frac{H}{16}}\times{\frac{W}{16}}\times D}$, and $T5\in\mathbb{R}^{{\frac{H}{32}}\times{\frac{W}{32}}\times D}$.}. 
It is noteworthy that each query in DETR concentrates on specific areas. Inspired by SOLO \cite{wang2020solo}, we first resize the restored feature maps $T3$-$T5$ to the shape of $S_{i}\times S_{i}\times D, i=1,2,3$. Then, we divide the resized features into $S_{i}\times S_{i}$ feature grids and flatten them to produce our location-guided queries $Q\in\mathbb{R}^{L\times D}, L=\sum_{i=1}^{3}S_{i}^{2}$. In this situation, the proposed location-guided queries can utilize learnable local features in different locations to optimize the initialization and efficiently aggregate the features in the camouflaged areas. Compared to the zero or random initialization \cite{carion2020end,zhu2020deformable}, this query generation strategy improves the efficiency of query iterations in the transformer decoder and accelerates the training convergence. For more discussion, please refer to \secref{sec:ablation}.

\begin{figure}[t!]
\centering
\includegraphics[width=.98\linewidth]{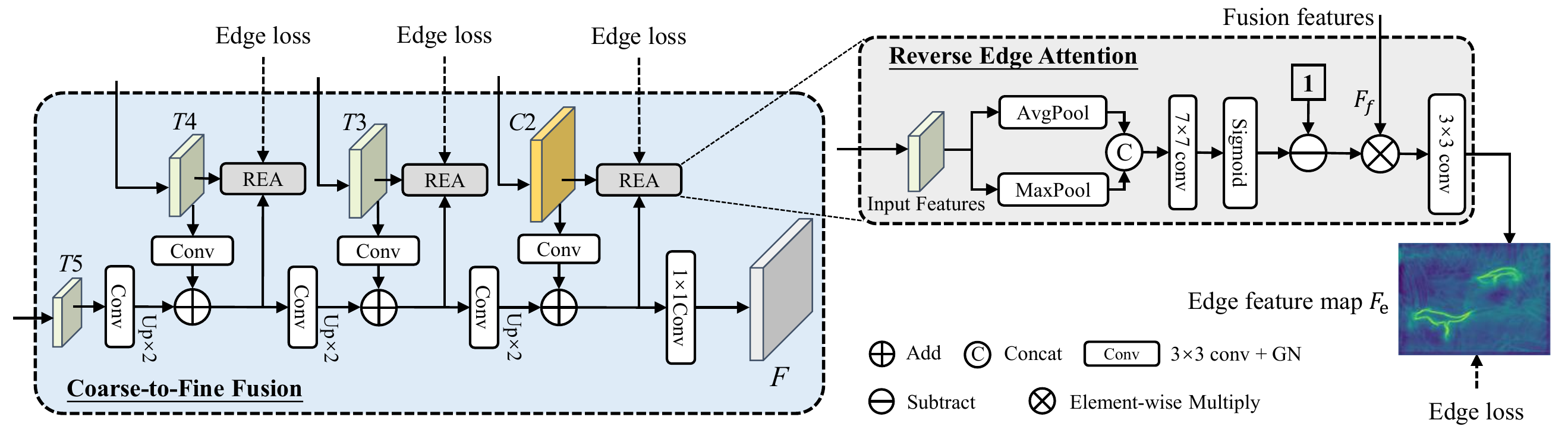}
\caption{Structure of our coarse-to-fine fusion.}
\label{fig:CFF}
\end{figure}

\noindent\textbf{LST Decoder.} 
The LST decoder is essential to interact with the global features produced by the LST encoder and location-guided queries for producing the instance-aware embeddings. Spatial position encoding is also added to our location-guided queries $Q_{L}$ and the encoder memory $X_{e}$. After that, they are fused by the deformable cross-attention layer. Unlike the general transformer decoder, we directly use cross-attention without self-attention because the proposed queries already contain learnable global features. BC-FFN is also employed after deformable attention operations, similar to the LST encoder. 
Given location-guided queries $Q_{L}$, the pipeline of our LST decoder is summarized as:
\begin{equation}
X_{d}=\textit{BC-FFN}(LN((Q_{L}+P_{s})+\textit{MDCAttn}((Q_{L}+P_{s}), (X_{e}+P_{m})))),
\end{equation}
where $P_{s}$ represents the position encoding based on the feature grids. \textit{MDCAttn} is denoted as the multi-head deformable cross-attention operation. $X_{d}$ is the output embeddings for instance-aware representation. Finally, $X_{d}$ is restored to feed into the following DCIN module (\secref{sec:DCIN}) for predicting masks.

\subsection{Coarse-to-Fine Fusion}\label{sec:CFF}
As a bottom-up transformer-based model, \ourmodel~strives to utilize multi-level global features output from the LST encoder to result in a shared mask feature representation. To merge diverse context information, we also fuse the low-level feature $C2$ from the CNN backbone as a complement to yield a unified high-resolution feature map $F\in\mathbb{R}^{{\frac{H}{4}}\times{\frac{W}{4}}\times D}$. The detailed structure of the proposed coarse-to-fine fusion (CFF) is shown in \figref{fig:CFF}. We take the multi-level features $C2, T3, T4$, and $T5$ as input for cascade fusion. Starting from $T5$ at 1/32 scale of input, a $3\times3$ convolution, GN, and $2\times$ bilinear upsampling are passed and added with the higher-resolution feature ($T4$ with 1/16 scale). After fusing $C2$ with a 1/4 scale, the feature proceeds through a $1\times1$ convolution, GN, and RELU operations to generate the mask feature $F$. Note that each input feature reduces the channels from 256 to 128 after the first convolution and then is increased to 256 channels at the final output.

Considering that the edge features of camouflage appear more challenging to capture, we design a reverse edge attention (REA) module embedded in CFF to supervise the edge features during the iterative process. 
Unlike the previous reverse attention~\cite{chen2018reverse,fan2020pranet}, our REA operates on the edge features rather than the predicted binary masks. In addition, the edge labels used for supervision are obtained by erosion of instance mask labels without any manual labeling. Inspired by the Convolutional Block Attention~\cite{woo2018cbam}, the input features are operated by both average-pooling $(AvgPool)$ and max-pooling $(MaxPool)$. Then, we concatenate and forward them to a $7\times7$ convolution and a sigmoid function. Afterward, we reverse the attention weight and apply them to the fusion feature $F_{f}$ by element-wise multiplication. 
Lastly, we use a $3\times3$ convolution to predict the edge feature. Assuming that the input feature is $Ti$, the whole process of each REA module can be formulated as follows:
\begin{equation}
F_{e}=Conv^{3}(F_{f} \otimes (1-Sigmoid(Conv^{7}([AvgPool(Ti); MaxPool(Ti)])))),
\end{equation}
where $Conv^{7}$ represents the $7\times7$ convolution layer, and $[;]$ denotes concatenation on the channel axis. In a word, the proposed CFF provides a shared mask feature $F$ to feed into the DCIN to predict the final camouflaged per-instance mask. 

\begin{figure}[t!]
\centering
\includegraphics[width=.98\linewidth]{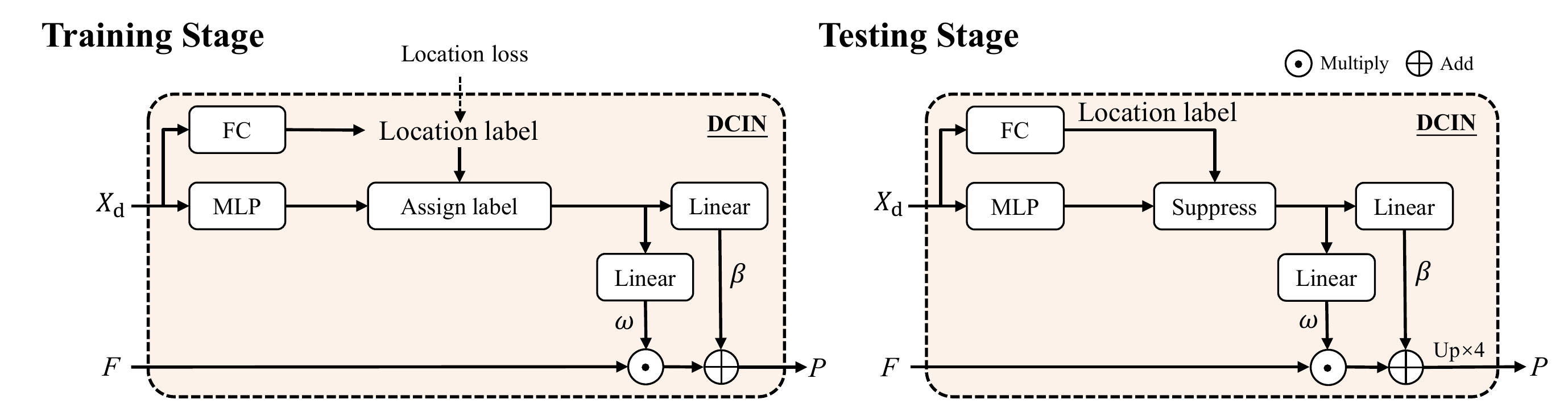}\caption{Structure of our dynamic camouflaged instance normalization.}
\label{fig:DCIN}
\end{figure}

\subsection{Dynamic Camouflaged Instance Normalization}\label{sec:DCIN}
Inspired by the instance normalization operation in the style transfer domain \cite{huang2017arbitrary, sofiiuk2019adaptis}, we introduce a dynamic camouflaged instance normalization (DCIN) to predict final masks. When DCIN receives the output embeddings ${X_{d}}\in\mathbb{R}^{S^{2}\times D}$ from the LST decoder, a fully-connected layer (FC) is employed to gain the location label. 
In parallel, a multi-layer perceptron (MLP) is used to gain the instance-aware parameters with a size of $D$ (\ie, 256). We assign positive and negative locations according to the ground truth in the training stage. The instance-aware parameters of the positive location are applied to generate the segmentation mask. In the testing stage, we utilize the confidence value of the location label to filter (See Suppress in \figref{fig:DCIN}) ineffective parameters (\eg, Threshold $>$ 0.5). Subsequently, two linear layers are operated on the filtered location-aware parameters to attain affine weights $\omega\in\mathbb{R}^{N\times D}$ and biases $\beta\in\mathbb{R}^{N\times 1}$, respectively. Finally, they are used together with the shared mask feature $F\in\mathbb{R}^{{\frac{H}{4}}\times{\frac{W}{4}}\times D}$ to predict the camouflaged instances, which can be described as:
\begin{equation}
P=U_{\times 4}(\omega F+\beta),
\end{equation}
where $P\in\mathbb{R}^{H\times W\times N}$ is the predicted masks. $N$ is the number of predicted instances. $U_{\times 4}$ is an upsampling operation by a factor of 4. In the end, the Matrix NMS~\cite{wang2020solov2} is applied to get the final instances.

\subsection{Loss Function}
During the training, the total loss function can be written as:
\begin{equation}
L_{total}=\lambda_{edge}L_{edge}+\lambda_{loc}L_{loc}+\lambda_{mask}L_{mask},
\end{equation}
where $L_{edge}$ is the edge loss to supervise edges from different levels in our CFF. The edge loss can be defined as $L_{edge}=\sum_{j=1}^{J}L_{dice}^{(j)}$, where $J$ represents the total number of levels of edge features for supervision, which can be seen in \figref{fig:CFF}. $\lambda_{edge}$ is the weight for edge loss that is set to 1 by default. Since the CIS task is category-agnostic, we use the confidence of the existence of camouflages in each location ($L_{loc}$) 
compared to the classification confidence in generic instance segmentation. In addition, $L_{loc}$ is implemented by Focal loss \cite{lin2017focal} and $L_{mask}$ is computed by Dice loss \cite{milletari2016v} for segmentation. $\lambda_{loc}$ and $\lambda_{mask}$ are set to 1 and 3 respectively to balance the total loss. 

\section{Experiments}
\subsection{Experimental Settings}
\noindent\textbf{Datasets}. As a brand-new challenging CIS task, few task-specific datasets so far. Comfortingly, Fan \etal~contributed a COD dataset \cite{fan2020camouflaged}, namely COD10K, which simultaneously provides high-quality instance-level annotations for training CIS models. Concretely, COD10K contains 3,040 camouflaged images with instance-level labels for training and 2,026 images for testing. Recently, Le \etal \cite{ltnghia-TIP2022} provided a larger CIS dataset called CAMO++, which includes a total of 5,500 samples with hierarchically pixel-wise annotation. Furthermore, Lyu \etal \cite{yunqiu_cod21} introduced a CIS test set with 4,121 images, called NC4K. 
We use the instance-level annotations in COD10K to train the proposed \ourmodel~and evaluate it on the COD10K and NC4K test set.

\noindent\textbf{Evaluation Metrics.} 
We adopt COCO-style evaluation metrics including AP$_{50}$, AP$_{75}$, and AP scores \cite{lin2014microsoft} to evaluate segmentation results. In contrast to the mAP metric in instance segmentation, each camouflaged instance detected from concealed regions is class-agnostic. Therefore, we only need to consider the existence of camouflaged instances while ignoring the mean value of the category.

\noindent\textbf{Technical Details.} 
Our \ourmodel~is implemented in PyTorch on a single RTX 3090 GPU and trained with Stochastic Gradient Descent. For fair comparisons, we adopt wisely used ResNet-50 backbone \cite{he2016deep} that is initialized from the pre-trained weights of ImageNet \cite{deng2009imagenet}. If not specially mentioned, other backbones used in our experiments are also pre-trained on ImageNet. During training, all our models are trained for 90K iterations (60 epochs) with a batch size of 2 and a base learning rate of $2.5e-4$ with a warm-up of 1K iterations. Then, the learning rate is divided by 10 at 60K and 80K, respectively. In addition, the weight decay is set to $10^{-4}$ and the momentum is 0.9. The input images are resized such that the size of the shortest side is from 480 to 800 while the longest side is at most 1,333. We also use the scale jittering augmentation for data augmentation.
In our LST, $S_{1}$, $S_{2}$, and $S_{3}$ are set to 36, 24, and 16, respectively. 
Note that the dimension of the features is kept at 256 throughout the whole process of BC-FFN. We embed a total of six encoder layers stacked sequentially. To reach better performance, we only repeat the LST decoder layer three times to aggregate camouflaged cues relevant to queries.

\subsection{Ablation Studies}\label{sec:ablation}
We conduct a series of ablation studies on the instance-level COD10K dataset \cite{fan2020camouflaged} to validate the effectiveness of our \ourmodel~and determine the hyper-parameters. The ablation experiments mainly consist of the following aspects: 
layers of encoder and decoder in LST,
number of multi-scale feature inputs, 
location-guided queries designs, 
feature fusion in the CFF module, 
backbone architecture, 
real-time settings,
and contributions of different components.

\begin{table*}[b!]
    \centering
    \scriptsize
    \renewcommand{\arraystretch}{0.1}
    \renewcommand{\tabcolsep}{6.4mm}
    \caption{Effect of the different number of encoder and decoder layers in our LST.}\label{tab:Number-LST}
    \begin{tabular}{c|c|ccc|c}
    Encoder     &     Decoder        &       AP              &           AP$_{50}$    &        AP$_{75}$      &       FPS       \\ \hline
    1                &       3                  &      37.0             &              68.0            &           35.4               &     \textbf{21.8}      \\ 
    3                &       1                  &      39.2             &              69.1            &           38.5               &     20.0      \\
    3                &       3                  &      39.4             &              70.2            &           39.3               &     18.8      \\ 
    3                &       6                  &      38.9             &              68.6            &           37.9               &     17.2      \\
    \rowcolor[RGB]{235,235,250}
    6                &       3                  &  \textbf{41.0}  &       \textbf{71.1}     &     40.8       &     14.5      \\
    6                &       6                  &      40.6            &               70.3           &            \textbf{41.2}          &     13.4      \\
    9                &       6                  &      40.7            &               70.6           &            40.4               &     11.3      \\
    \end{tabular}
\end{table*}

\noindent\textbf{Layers of Encoder and Decoder in LST.} 
The depth of the transformer is a key factor influencing the performance and efficiency of transformer-based models. We attempt multiple combinations of different numbers of encoder and decoder layers in our LST to optimize the performance of \ourmodel. 
As shown in \tabref{tab:Number-LST}, the first three rows indicate that three layers are insufficient to maximize the performance of \ourmodel. 
In addition, we observe that LST is more sensitive to the encoder than the decoder. The highest value of AP is reached when the number of encoder and decoder layers are 6 and 3, respectively. 
When more layers are added, the accuracy is not further improved, and the inference time drops to under 14$fps$. As a result, we adopt 6 encoder layers and 3 decoder layers as our default setting to balance performance and efficiency.

\begin{table*}[t!]
	    \centering
	    \scriptsize
	    \renewcommand{\arraystretch}{0.1}
        \renewcommand{\tabcolsep}{4.5mm}
	    \caption{Ablations for different combinations of multi-scale features input to our LST.}\label{tab:CNN-inputs}
	    \begin{tabular}{c|c|ccc|r|r}
	    Scales     &   Number   &   AP    &  AP$_{50}$  &  AP$_{75}$   &    Params   &   Memory        \\ \hline
	    \rowcolor[RGB]{235,235,250}
	    $C3$-$C5$     &       3         &  \textbf{41.0}   &   \textbf{71.1}   &   \textbf{40.8}     &  \textbf{46.58M}   &  \textbf{6.4G}       \\ 
	    $C2$-$C5$     &       4         &  39.9   &        70.5        &    38.7            &     46.80M        &    9.3G              \\
	    $C3$-$C6$     &       4         &  40.8   &        70.6        &    40.9            &      47.39M        &    6.7G              \\ 
	    $C2$-$C6$     &       5         &  40.2   &        69.9        &    40.3           &      47.62M        &    17.7G            \\
	    \end{tabular}
\end{table*}

\noindent\textbf{Number of Multi-scale Feature Inputs.} 
We utilize multi-level features extracted from the ResNet-50 as input of our LST. 
To more accurately capture camouflages at different scales while maintaining model efficiency,
we combine different numbers of features in the backbone, including $C3$-$C5$, $C2$-$C5$, $C3$-$C6$, and $C2$-$C6$. In \tabref{tab:CNN-inputs}, we observe that the combination of $C3$-$C5$ achieves a strong performance with the lowest number of parameters and training memory. 

\noindent\textbf{Location-Guided Queries Designs.}
Object queries are essential in the transformer architecture for dense prediction tasks. To validate the effectiveness of our location-guided queries, we compare two typical object query designs, including zero-initialized in vanilla DETR~\cite{carion2020end} and 
learnable input embeddings in deformable DETR~\cite{zhu2020deformable}. We set the number of queries uniformly to the default number of multi-scale feature grids for fair comparisons. Other settings in the proposed \ourmodel~remain unchanged.
In a nutshell, object queries in transformer decoder comprise two parts: query features and query position embeddings. In the vanilla DETR, an all-zero matrix added by a set of learnable position embeddings is taken as object queries through the decoder to generate the corresponding output embeddings. In contrast, the deformable DETR is directly initialized by learnable embeddings as query features and is coupled with learnable position embeddings.
As seen in \tabref{tab:Query}, our location-guided queries are significantly superior to other query designs. It illustrates that inserting supervised global features in queries is crucial to regressing different camouflage cues and locating instances efficiently. Furthermore, we compare the learning ability of three strategies. We find that our location-guided queries scheme has a faster convergence rate at the early training stage, and the final convergence is also better than the other two models. 
It also demonstrates that location-guided queries are 
efficient to exploit global features to capture camouflaged information at different locations by cross-attention mechanism.

\begin{table*}[!t]
	    \centering
	    \scriptsize
	    \renewcommand{\arraystretch}{0.1}
        \renewcommand{\tabcolsep}{5.7mm}
	    \caption{Comparison of different query designs on the proposed \ourmodel.}\label{tab:Query}
	    \begin{tabular}{l|lll}
	    Queries                                                                          &       AP                    &      AP$_{50}$    &      AP$_{75}$         \\ \hline
	    Zero-Initialized~\cite{carion2020end}                          &       34.7                  &      64.1             &             33.1                 \\ 
	    Learnable Embeddings~\cite{zhu2020deformable}       &       35.0                  &      64.8             &             33.2                   \\
	    \rowcolor[RGB]{235,235,250}
	    Location-Guided Queries (\textbf{Ours})                                   &       \textbf{41.0}$_{~+6.0}$                  &      \textbf{71.1}$_{~+6.3}$            &              \textbf{40.8}$_{~+7.6}$              \\ 
	    \end{tabular}
\end{table*}

\begin{table*}[t!]
	    \centering
	    \scriptsize
	    \renewcommand{\arraystretch}{0.1}
        \renewcommand{\tabcolsep}{10.6mm}
	    \caption{Comparison of different feature combinations input to our CFF module.}\label{tab:CFF}
	    \begin{tabular}{c|ccc}
	    Features                 &          AP             &      AP$_{50}$    &      AP$_{75}$         \\ \hline
	    Single $T2$               &         38.0            &      69.2               &             36.8               \\ 
	    $C2, C3, C4, C5$      &         35.4            &      64.3               &             34.6               \\
	    $C2, C3, C4, T5$      &         40.0            &      69.7               &             40.1                \\ 
	    $C2, C3, T4, T5$      &         39.5            &      69.9               &             39.0                 \\
   	    $T2, T3, T4, T5$      &         40.0            &       70.1              &             40.0                 \\
   	    \rowcolor[RGB]{235,235,250}
	    $C2, T3, T4, T5$      &   \textbf{41.0}   &     \textbf{71.1}  &     \textbf{40.8}           \\  
	    \end{tabular}
\end{table*}

\begin{table*}[t!]
	    \centering
	    \scriptsize
	    \renewcommand{\arraystretch}{0.1}
        \renewcommand{\tabcolsep}{6.6mm}
	    \caption{Performance of \ourmodel~with different backbone networks.}\label{tab:backbone}
	    \begin{tabular}{l|ccc|c}
	    Backbone Networks                                                  &          AP             &      AP$_{50}$    &      AP$_{75}$      &        FPS                 \\ \hline
	    \rowcolor[RGB]{235,235,250}
	    ResNet-50 \cite{he2016deep} (\textbf{Default})                 &         41.0            &        71.1             &             40.8           &    \textbf{14.5}       \\ 
	    ResNet-101 \cite{he2016deep}                &         42.0            &        71.3             &             42.8           &        12.9                 \\
	    PVTv2-B2-Li  \cite{wang2021pvtv2}     &    47.2     &     74.9      &  \textbf{49.8}       &        13.2                 \\
	    Swin-T \cite{liu2021swin}    &  \textbf{47.7}     &  \textbf{78.6}     &  49.3       &        12.6                 \\
	    \end{tabular}
\end{table*}

\begin{figure}[b!]
\centering
\includegraphics[width=.98\linewidth]{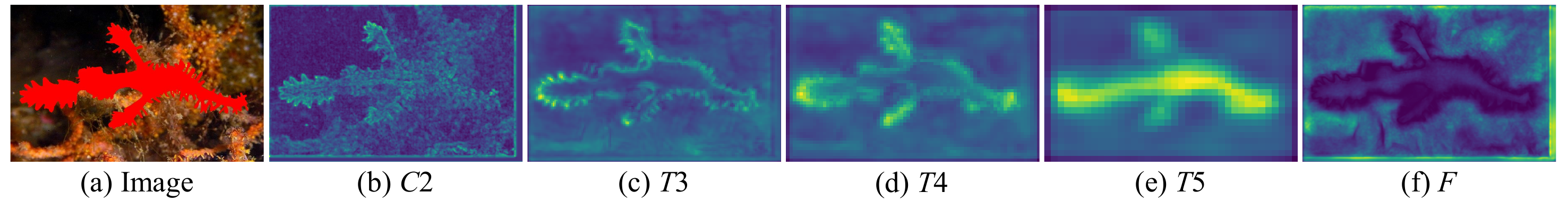}
\caption{Visualizations of feature maps. (a) The input image overlapped with ground-truth; (b)-(e) is the input features of CFF produced from the CNN backbone and our LST encoder; (f) is the mask feature $F$ output from CFF.}
\label{fig:CFF-Feature}
\end{figure}

\noindent\textbf{Feature Fusion in CFF.}
In the proposed CFF module, the multi-scale input features directly impact the quality of the mask feature $F$ operated by fusing. To explore the optimal fusion scheme from ResNet-50 and LST encoder, we conduct different combinations in \tabref{tab:CFF}. Using only a single-scale feature $T2$ without 
multi-scale fusion is not promising. The result of the 2$^{nd}$ row illustrates that fusing only the backbone features is inefficient. In the end, we attain the optimal results by feeding $C2, T3, T4$, and $T5$ into the CFF module. It can be explained that the features from our LST encoder have more detailed global information. Furthermore, the feature $C2$ is also required to provide some low-level features as a supplement. In addition, we visualize the features at each scale input to the CFF module and the mask feature $F$ in \figref{fig:CFF-Feature}. 

\noindent\textbf{Backbone Networks.}
In this experiment, we use different backbone networks \ie, ResNet-50~\cite{he2016deep}, ResNet-101 \cite{he2016deep}, PVTv2-B2-Li~\cite{wang2021pvtv2}, and Swin-T~\cite{liu2021swin}, to train our \ourmodel. All of them are pre-trained on ImageNet~\cite{deng2009imagenet}. 
From \tabref{tab:backbone}, we observe that \ourmodel~only with the ResNet-50 can achieve 41\% AP scores. Moreover, using a more powerful backbone can further stimulate the potential of our method that further improves to 47.7\% AP.

\noindent\textbf{Real-Time Settings.} 
To improve the application value of \ourmodel, we provide a real-time version named \ourmodel-550. Concretely, we resize the input shorter side to 550 while reducing the LST encoder layers to 3. 
As shown in \tabref{tab:Time}, despite the value of AP dropping to 36.0\%, the inference time is increased to 25.8$fps$, and the number of parameters and FLOPs is also significantly improved. We hope \ourmodel-550 can be extended to more real-life scenarios. 

\begin{table*}[t!]
	    \centering
	    \scriptsize
	    \renewcommand{\arraystretch}{0.1}
        \renewcommand{\tabcolsep}{3.1mm}
	    \caption{
	    Performance of \ourmodel~with real-time setting. 
	    }\label{tab:Time}
	    \begin{tabular}{l|c|ccc|c|c}
	    Models                            &     Backbones        &       AP              &           AP$_{50}$    &        FPS          &       Params         &    FLOPs        \\ \hline
	    \rowcolor[RGB]{235,235,250}
	    \ourmodel~(\textbf{Default})              &      ResNet-50        &      \textbf{41.0}             &              \textbf{71.1}            &        14.5          &         46.6M           &    324.7G        \\ 
	    \ourmodel-550                     &      ResNet-50        &      36.0             &              65.3            &         \textbf{25.8}         &         \textbf{42.4M}           &    \textbf{138.7G}        \\
	    \end{tabular}
\end{table*}

\begin{table*}[t!]
	    \centering
	    \scriptsize
	    \renewcommand{\arraystretch}{0.1}
        \renewcommand{\tabcolsep}{4.0mm}
	    \caption{Ablation studies for different components in our \ourmodel. }\label{tab:Components}
	    \begin{tabular}{c|c|c|c|c|ccc}
	    Encoder        &         LGQ          &         BC-FFN    &         CFF         &        REA         &        AP            &   AP$_{50}$      &  AP$_{75}$          \\ \hline
	                         &    \checkmark     &    \checkmark    &    \checkmark   &   \checkmark   &       33.7           &        63.4            &      32.0                   \\ 
	    \checkmark  &                            &    \checkmark    &    \checkmark   &   \checkmark   &       34.7           &        64.1            &      33.1                   \\
	    \checkmark  &    \checkmark     &                           &    \checkmark   &   \checkmark   &       37.2           &        67.3            &      35.8                   \\ 
	    \checkmark  &    \checkmark     &    \checkmark    &                          &                         &       38.0           &        69.2            &      36.8                   \\
	    \checkmark  &    \checkmark     &    \checkmark    &    \checkmark   &                         &       39.3           &        69.7            &      38.5                   \\
	    \rowcolor[RGB]{235,235,250}
	    \checkmark  &    \checkmark     &    \checkmark    &    \checkmark   &   \checkmark   &  \textbf{41.0}  &   \textbf{71.1}   &   \textbf{40.8}        \\  
	    \end{tabular}
\end{table*}

\begin{table*}[!t]
    \centering
    \scriptsize
    \renewcommand{\arraystretch}{0.1}
    \renewcommand{\tabcolsep}{1.0mm}
    \caption{Quantitative comparisons with 11 representative methods.
    }\label{tab:SOTA}
\begin{tabular}{c|l|l|c|c|ccc|ccc}
&\multirow{2}{*}{Methods} & \multirow{2}{*}{Backbones} & \multirow{2}{*}{Params} &\multirow{2}{*}{FLOPs}   & \multicolumn{3}{c|}{COD10K-Test} & \multicolumn{3}{c}{NC4K-Test} \\ 
\cline{6-11} 
&                        &            &            &                 &   AP     &  AP$_{50}$    & AP$_{75}$    & AP   & AP$_{50}$ & AP$_{75}$   \\ 
\hline
\multirow{12}{*}{\begin{sideways}Two-Stage\end{sideways}}
& Mask R-CNN \cite{he2017mask}                  & ResNet-50    &    43.9M   &   186.3G      &          25.0       &       55.5                 &          20.4              &          27.7     &   58.6         &      22.7             \\
&Mask R-CNN \cite{he2017mask}                  & ResNet-101    &    62.9M     &    254.5G       &         28.7       &       60.1                 &          25.7              &       36.1      &      68.9         &        33.5               \\ 
&MS R-CNN \cite{huang2019mask}                     & ResNet-50      &   60.0M   &      198.5G    &         30.1         &              57.2             &             28.7              &        31.0     &    58.7                &       29.4            \\
&MS R-CNN \cite{huang2019mask}                     & ResNet-101    &  79.0M     &    251.1G      &          33.3       &         61.0          &        32.9             &        35.7          &    63.4              &     34.7            \\ 
&Cascade R-CNN \cite{cai2019cascade}   & ResNet-50       &   71.7M      &   334.1G     &        25.3       &            56.1               &          21.3            &             29.5       &     60.8       &          24.8               \\
&Cascade R-CNN \cite{cai2019cascade}   & ResNet-101     &    90.7M    &    386.7G      &           29.5        &           61.0           &            25.9             &             34.6       &     66.3       &         31.5             \\ 
&HTC \cite{chen2019hybrid}                               & ResNet-50        &    76.9M    &    331.7G     &         28.1        &             56.3              &           25.1                 &       29.8      &          59.0         &        26.6     \\
&HTC \cite{chen2019hybrid}                              & ResNet-101      &  95.9M    &   384.3G       &         30.9       &          61.0                 &             28.7               &             34.2       &          64.5       &        31.6  \\ 
&BlendMask \cite{chen2020blendmask}                    & ResNet-50         &    35.8M     &       233.8G     &         28.2     &       56.4                &           25.2               &        27.7    &    56.7            &       24.2                \\
&BlendMask \cite{chen2020blendmask}                    & ResNet-101       &   54.7M     &    302.8G      &          31.2     &       60.0                   &           28.9               &       31.4    &     61.2           &       28.8     \\
&Mask Transfiner \cite{ke2022mask}    & ResNet-50         &    44.3M    &   \textbf{185.1}G       &       28.7       &        56.3          &        26.4                &   29.4      &     56.7       &       27.2            \\
&Mask Transfiner \cite{ke2022mask}    & ResNet-101         &    63.3M     &    253.7G      &      31.2     &     60.7          &   29.8                    &   34.0       &       63.1     &         32.6         \\
\hline
\multirow{14}{*}{\begin{sideways}One-Stage\end{sideways}}
&YOLACT \cite{bolya2019yolact}                      & ResNet-50         &   -   &     -    &          24.3      &       53.3                &           19.7               &       32.1    &  65.3            &     27.9            \\
&YOLACT \cite{bolya2019yolact}                      & ResNet-101       &   -     &   -   &          29.0          &       60.1                &           25.3               &      37.8     &    70.6            &      35.6        \\ 
&CondInst \cite{tian2020conditional}                       & ResNet-50          &   \textbf{34.1}M      &    200.1G     &         30.6       &        63.6               &            26.1              &     33.4       &        67.4               &            29.4                 \\
&CondInst \cite{tian2020conditional}                       & ResNet-101        &   53.1M      &     269.1G     &          34.3       &        67.9               &            31.6              &         38.0    &         71.1       &       35.6               \\ 
&QueryInst \cite{fang2021instances}                     & ResNet-50          &    -     &    -     &          28.5      &         60.1          &          23.1              &     33.0      &      66.7      &      29.4              \\
&QueryInst \cite{fang2021instances}                     & ResNet-101        &    -    &      -   &         32.5        &            65.1             &                28.6              &       38.7      &         72.1         &       37.6        \\ 
&SOTR \cite{guo2021sotr}                          & ResNet-50          &    63.1M        &    476.7G      &          27.9      &        58.7               &            24.1               &       29.3     &   61.0        &         25.6       \\
&SOTR \cite{guo2021sotr}                           & ResNet-101        &    82.1M      &    549.6G      &      32.0       &       63.6               &            29.2               &          34.3      &   65.7              &   32.4             \\ 
&SOLOv2 \cite{wang2020solov2}                      & ResNet-50          &    46.2M          &     318.7G     &          32.5       &       63.2               &            29.9              &            34.4      &   65.9            &        31.9             \\
&SOLOv2 \cite{wang2020solov2}                      & ResNet-101        &    65.1M     &    394.6G      &        35.2       &       65.7               &            33.4              &       37.8     &    69.2          &          36.1          \\ 
\rowcolor[RGB]{235,235,250}
&\textbf{\ourmodel~(Ours)}                             & ResNet-50          &       46.6M    &     324.7G    &      \underline{41.0}      &    \underline{71.1}      &    \underline{40.8}      &    \underline{42.5}    &   \underline{72.5}    &    \underline{42.3}             \\
\rowcolor[RGB]{235,235,250}
&\textbf{\ourmodel~(Ours)}                              & ResNet-101       &   65.5M     &    398.2G      &      \textbf{42.0}   &  \textbf{71.3}  &   \textbf{42.8}           &  \textbf{44.4}    &    \textbf{73.7}      &    \textbf{45.1}        \\
\end{tabular}
\end{table*}

\noindent\textbf{Contributions of Different Components.}
We conduct extensive ablation studies on the COD10K test set \cite{fan2020camouflaged}, including LST encoder (Encoder), location-guided queries (LGQ), blend-convolution FFN (BC-FFN), coarse-to-fine fusion (CFF) and reverse edge attention (REA). We adopt a control variable manner, where we only ablate the current module while keeping the other parts as default settings. When validating the LGQ, we use the learnable embeddings~\cite{zhu2020deformable} as an alternative. Similarly, BC-FFN is replaced by the vanilla FFN~\cite{vaswani2017attention}. For the CFF module, we directly use a single-scale feature $T2$ as the output of CFF. As shown in \tabref{tab:Components}, without Encoder, the value of AP directly dropped by about 7\%. It indicates that the LST encoder is essential to extract high-level global features. Furthermore, the 2$^{nd}$ row validates the effectiveness 
of LGQ design again.
Note that BC-FFN plays a vital role in the encoder and decoder of LST 
because $3\times3$ convolution layers can strengthen the local correlation of global features from self-attention. Moreover, the CFF efficiently fuses multi-scale features and enhances the edges of camouflaged instances by embedding REA. By integrating all modules, \ourmodel~reaches best performance.

\subsection{Comparisons with Cutting-Edge Methods}
We compare our \ourmodel~with several famous instance segmentation models (\ie, two-stage and one-stage models) retrained on the instance-level COD10K~\cite{fan2020camouflaged} dataset. 
We uniformly adopt official codes for fair comparisons to train each model and evaluate them on the COD10K and NC4K~\cite{yunqiu_cod21} test set. 
In addition, we also show the results based on different backbones \ie, ResNet-50, ResNet-101, with the ImageNet~\cite{deng2009imagenet} pre-trained weights.

\noindent\textbf{Quantitative Comparisons.}
As shown in \tabref{tab:SOTA}, although the CIS task is challenging, our \ourmodel~still performs favorably against other competitors across all metrics. 
In particular, the AP score of \ourmodel~is higher than that of the second-ranked SOLOv2 \cite{wang2020solov2} by a large margin ($\sim$8.5\%) using ResNet-50.
The desirable result should be attributed to our LST because it provided higher-level global features and interacted with camouflage clues in different locations in the LST decoder.
By leveraging a more powerful backbone, \ie, Swin-T, \ourmodel~can continue to boost the performance to 47.7\% AP (\tabref{tab:backbone}). 
According to the parameters and FLOPs in \tabref{tab:SOTA}, it also demonstrates that our \ourmodel~achieves better performance without adding extra parameters.

\begin{figure}[t!]
\centering
\includegraphics[width=.8\linewidth]{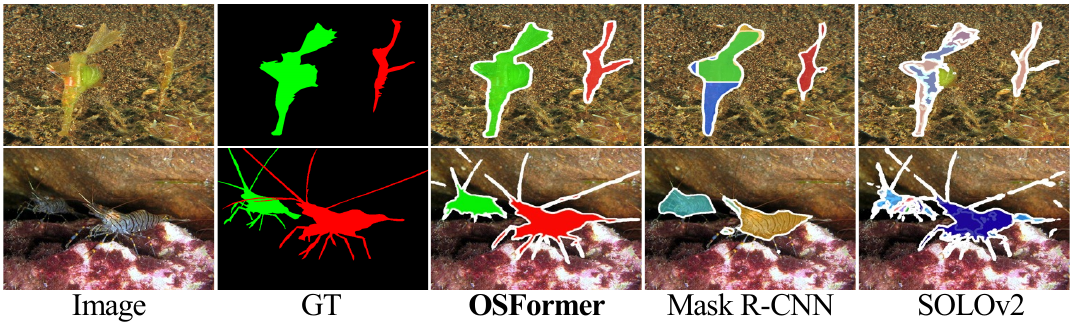}
\caption{Qualitative comparison of \ourmodel~with Mask R-CNN and SOLOv2.}\label{fig:Results}
\end{figure}

\noindent\textbf{Qualitative Comparisons.} 
To validate the effectiveness of \ourmodel, we also exhibit two representative visualization results in \figref{fig:Results}. 
Specifically, the top sample suggests that \ourmodel~can comfortably distinguish camouflages in the case of multiple instances. The bottom result shows that our method excels at capturing slender boundaries, which can be attributed to the enhancement of edge features by our REA module.
Overall, compared to the visualization results from other famous methods, \ourmodel~has the capability to overcome more challenging cases and achieve good performance.

\section{Conclusion}
We contributed a novel location-aware one-stage transformer framework, called \textbf{\ourmodel}, for camouflaged instance segmentation (CIS). \ourmodel~embraces an efficient location-sensing transformer to capture global features and dynamically regress the location and body of the camouflaged instances. 
As the first one-stage bottom-up CIS framework, we further designed a coarse-to-fine fusion to integrate multi-scale features and highlight the edge of camouflages to produce the global features. 
Extensive experimental results show that \ourmodel~performs favorably against all other well-known models. Furthermore, \ourmodel~requires only about 3,000 images to train, and it converges rapidly. 
It is easily and flexibly extended to other downstream vision tasks with smaller training samples. 


\bibliographystyle{splncs04}
\bibliography{egbib}

\begin{thebibliography}{10}
\providecommand{\url}[1]{\texttt{#1}}
\providecommand{\urlprefix}{URL }
\providecommand{\doi}[1]{https://doi.org/#1}

\bibitem{bai2017deep}
Bai, M., Urtasun, R.: Deep watershed transform for instance segmentation. In:
  IEEE CVPR (2017)

\bibitem{bhajantri2006camouflage}
Bhajantri, N.U., Nagabhushan, P.: Camouflage defect identification: a novel
  approach. In: IEEE ICIT (2006)

\bibitem{bolya2019yolact}
Bolya, D., Zhou, C., Xiao, F., Lee, Y.J.: Yolact: Real-time instance
  segmentation. In: IEEE CVPR (2019)

\bibitem{cai2019cascade}
Cai, Z., Vasconcelos, N.: Cascade r-cnn: high quality object detection and
  instance segmentation. IEEE TPAMI  \textbf{43}(5),  1483--1498 (2019)

\bibitem{carion2020end}
Carion, N., Massa, F., Synnaeve, G., Usunier, N., Kirillov, A., Zagoruyko, S.:
  End-to-end object detection with transformers. In: ECCV (2020)

\bibitem{chen2020blendmask}
Chen, H., Sun, K., Tian, Z., Shen, C., Huang, Y., Yan, Y.: Blendmask: Top-down
  meets bottom-up for instance segmentation. In: IEEE CVPR (2020)

\bibitem{chen2019hybrid}
Chen, K., Pang, J., Wang, J., Xiong, Y., Li, X., Sun, S., Feng, W., Liu, Z.,
  Shi, J., Ouyang, W., et~al.: Hybrid task cascade for instance segmentation.
  In: IEEE CVPR (2019)

\bibitem{chen2017deeplab}
Chen, L.C., Papandreou, G., Kokkinos, I., Murphy, K., Yuille, A.L.: Deeplab:
  Semantic image segmentation with deep convolutional nets, atrous convolution,
  and fully connected crfs. IEEE TPAMI  \textbf{40}(4),  834--848 (2017)

\bibitem{chen2018reverse}
Chen, S., Tan, X., Wang, B., Hu, X.: Reverse attention for salient object
  detection. In: ECCV (2018)

\bibitem{chu2010camouflage}
Chu, H.K., Hsu, W.H., Mitra, N.J., Cohen-Or, D., Wong, T.T., Lee, T.Y.:
  Camouflage images. ACM TOG  \textbf{29}(4),  51--1 (2010)

\bibitem{cuthill2019camouflage}
Cuthill, I.: Camouflage. JOZ  \textbf{308}(2),  75--92 (2019)

\bibitem{dai2021dynamic}
Dai, X., Chen, Y., Yang, J., Zhang, P., Yuan, L., Zhang, L.: Dynamic detr:
  End-to-end object detection with dynamic attention. In: IEEE CVPR (2021)

\bibitem{dai2021up}
Dai, Z., Cai, B., Lin, Y., Chen, J.: Up-detr: Unsupervised pre-training for
  object detection with transformers. In: IEEE CVPR (2021)

\bibitem{deng2009imagenet}
Deng, J., Dong, W., Socher, R., Li, L.J., Li, K., Fei-Fei, L.: Imagenet: A
  large-scale hierarchical image database. In: IEEE CVPR (2009)

\bibitem{dosovitskiy2020image}
Dosovitskiy, A., Beyer, L., Kolesnikov, A., Weissenborn, D., Zhai, X.,
  Unterthiner, T., Dehghani, M., Minderer, M., Heigold, G., Gelly, S., et~al.:
  An image is worth 16x16 words: Transformers for image recognition at scale.
  In: ICLR (2021)

\bibitem{fan2020camouflaged}
Fan, D.P., Ji, G.P., Sun, G., Cheng, M.M., Shen, J., Shao, L.: Camouflaged
  object detection. In: IEEE CVPR (2020)

\bibitem{fan2020pranet}
Fan, D.P., Ji, G.P., Zhou, T., Chen, G., Fu, H., Shen, J., Shao, L.: Pranet:
  Parallel reverse attention network for polyp segmentation. In: MICCAI (2020)

\bibitem{fan2020inf}
Fan, D.P., Zhou, T., Ji, G.P., Zhou, Y., Chen, G., Fu, H., Shen, J., Shao, L.:
  Inf-net: Automatic covid-19 lung infection segmentation from ct images. IEEE
  TMI  \textbf{39}(8),  2626--2637 (2020)

\bibitem{fang2021instances}
Fang, Y., Yang, S., Wang, X., Li, Y., Fang, C., Shan, Y., Feng, B., Liu, W.:
  Instances as queries. In: IEEE CVPR (2021)

\bibitem{fennell2021camouflage}
Fennell, J.G., Talas, L., Baddeley, R.J., Cuthill, I.C., Scott-Samuel, N.E.:
  The camouflage machine: Optimizing protective coloration using deep learning
  with genetic algorithms. Evolution  \textbf{75}(3),  614--624 (2021)

\bibitem{gao2019ssap}
Gao, N., Shan, Y., Wang, Y., Zhao, X., Yu, Y., Yang, M., Huang, K.: Ssap:
  Single-shot instance segmentation with affinity pyramid. In: IEEE CVPR (2019)

\bibitem{guo2021sotr}
Guo, R., Niu, D., Qu, L., Li, Z.: Sotr: Segmenting objects with transformers.
  In: IEEE ICCV (2021)

\bibitem{he2017mask}
He, K., Gkioxari, G., Doll{\'a}r, P., Girshick, R.: Mask r-cnn. In: IEEE ICCV
  (2017)

\bibitem{he2016deep}
He, K., Zhang, X., Ren, S., Sun, J.: Deep residual learning for image
  recognition. In: IEEE CVPR (2016)

\bibitem{huang2017arbitrary}
Huang, X., Belongie, S.: Arbitrary style transfer in real-time with adaptive
  instance normalization. In: IEEE ICCV (2017)

\bibitem{huang2019mask}
Huang, Z., Huang, L., Gong, Y., Huang, C., Wang, X.: Mask scoring r-cnn. In:
  IEEE CVPR (2019)

\bibitem{huerta2007improving}
Huerta, I., Rowe, D., Mozerov, M., Gonz{\`a}lez, J.: Improving background
  subtraction based on a casuistry of colour-motion segmentation problems. In:
  Iberian PRIA (2007)

\bibitem{ji2021progressively}
Ji, G.P., Chou, Y.C., Fan, D.P., Chen, G., Fu, H., Jha, D., Shao, L.:
  Progressively normalized self-attention network for video polyp segmentation.
  In: MICCAI (2021)

\bibitem{ke2022mask}
Ke, L., Danelljan, M., Li, X., Tai, Y.W., Tang, C.K., Yu, F.: Mask transfiner
  for high-quality instance segmentation. In: IEEE CVPR (2022)

\bibitem{ltnghia-TIP2022}
Le, T.N., Cao, Y., Nguyen, T.C., Le, M.Q., Nguyen, K.D., Do, T.T., Tran, M.T.,
  Nguyen, T.V.: Camouflaged instance segmentation in-the-wild: Dataset, method,
  and benchmark suite. IEEE TIP  \textbf{31},  287--300 (2022)

\bibitem{le2019anabranch}
Le, T.N., Nguyen, T.V., Nie, Z., Tran, M.T., Sugimoto, A.: Anabranch network
  for camouflaged object segmentation. CVIU  \textbf{184},  45--56 (2019)

\bibitem{lin2017focal}
Lin, T.Y., Goyal, P., Girshick, R., He, K., Doll{\'a}r, P.: Focal loss for
  dense object detection. In: IEEE ICCV (2017)

\bibitem{lin2014microsoft}
Lin, T.Y., Maire, M., Belongie, S., Hays, J., Perona, P., Ramanan, D.,
  Doll{\'a}r, P., Zitnick, C.L.: Microsoft coco: Common objects in context. In:
  ECCV (2014)

\bibitem{liu2017sgn}
Liu, S., Jia, J., Fidler, S., Urtasun, R.: Sgn: Sequential grouping networks
  for instance segmentation. In: IEEE ICCV (2017)

\bibitem{liu2018path}
Liu, S., Qi, L., Qin, H., Shi, J., Jia, J.: Path aggregation network for
  instance segmentation. In: IEEE CVPR (2018)

\bibitem{liu2021swin}
Liu, Z., Lin, Y., Cao, Y., Hu, H., Wei, Y., Zhang, Z., Lin, S., Guo, B.: Swin
  transformer: Hierarchical vision transformer using shifted windows. In: IEEE
  CVPR (2021)

\bibitem{yunqiu_cod21}
Lyu, Y., Zhang, J., Dai, Y., Li, A., Liu, B., Barnes, N., Fan, D.P.:
  Simultaneously localize, segment and rank the camouflaged objects. In: IEEE
  CVPR (2021)

\bibitem{matthews2015human}
Matthews, O., Liggins, E., Volonakis, T., Scott-Samuel, N., Baddeley, R.,
  Cuthill, I.: Human visual search performance for camouflaged targets. Journal
  of Vision  \textbf{15}(12),  1164--1164 (2015)

\bibitem{mei2021camouflaged}
Mei, H., Ji, G.P., Wei, Z., Yang, X., Wei, X., Fan, D.P.: Camouflaged object
  segmentation with distraction mining. In: IEEE CVPR (2021)

\bibitem{milletari2016v}
Milletari, F., Navab, N., Ahmadi, S.A.: V-net: Fully convolutional neural
  networks for volumetric medical image segmentation. In: IEEE 3DV (2016)

\bibitem{mondal2020camouflaged}
Mondal, A.: Camouflaged object detection and tracking: A survey. IJIG
  \textbf{20}(04),  2050028 (2020)

\bibitem{redmon2016you}
Redmon, J., Divvala, S., Girshick, R., Farhadi, A.: You only look once:
  Unified, real-time object detection. In: IEEE CVPR (2016)

\bibitem{ren2021deep}
Ren, J., Hu, X., Zhu, L., Xu, X., Xu, Y., Wang, W., Deng, Z., Heng, P.A.: Deep
  texture-aware features for camouflaged object detection. IEEE TCSVT  (2021)

\bibitem{ren2015faster}
Ren, S., He, K., Girshick, R., Sun, J.: Faster r-cnn: Towards real-time object
  detection with region proposal networks. In: NeurIPS (2015)

\bibitem{sandon1990simulating}
Sandon, P.A.: Simulating visual attention. Journal of Cognitive Neuroscience
  \textbf{2}(3),  213--231 (1990)

\bibitem{sofiiuk2019adaptis}
Sofiiuk, K., Barinova, O., Konushin, A.: Adaptis: Adaptive instance selection
  network. In: IEEE CVPR (2019)

\bibitem{song2010new}
Song, L., Geng, W.: A new camouflage texture evaluation method based on wssim
  and nature image features. In: ICMT (2010)

\bibitem{stevens2009animal}
Stevens, M., Merilaita, S.: Animal camouflage: current issues and new
  perspectives. PTRS B: BS  \textbf{364}(1516),  423--427 (2009)

\bibitem{tian2020conditional}
Tian, Z., Shen, C., Chen, H.: Conditional convolutions for instance
  segmentation. In: ECCV (2020)

\bibitem{tian2019fcos}
Tian, Z., Shen, C., Chen, H., He, T.: Fcos: Fully convolutional one-stage
  object detection. In: IEEE ICCV (2019)

\bibitem{troscianko2021variable}
Troscianko, J., Nokelainen, O., Skelhorn, J., Stevens, M.: Variable crab
  camouflage patterns defeat search image formation. Communications biology
  \textbf{4}(1), ~1--9 (2021)

\bibitem{vaswani2017attention}
Vaswani, A., Shazeer, N., Parmar, N., Uszkoreit, J., Jones, L., Gomez, A.N.,
  Kaiser, {\L}., Polosukhin, I.: Attention is all you need. In: NeurIPS (2017)

\bibitem{wang2021max}
Wang, H., Zhu, Y., Adam, H., Yuille, A., Chen, L.C.: Max-deeplab: End-to-end
  panoptic segmentation with mask transformers. In: IEEE CVPR (2021)

\bibitem{wang2021pyramid}
Wang, W., Xie, E., Li, X., Fan, D.P., Song, K., Liang, D., Lu, T., Luo, P.,
  Shao, L.: Pyramid vision transformer: A versatile backbone for dense
  prediction without convolutions. In: IEEE CVPR (2021)

\bibitem{wang2021pvtv2}
Wang, W., Xie, E., Li, X., Fan, D.P., Song, K., Liang, D., Lu, T., Luo, P.,
  Shao, L.: Pvtv2: Improved baselines with pyramid vision transformer. CVMJ
  (2022)

\bibitem{wang2020solo}
Wang, X., Kong, T., Shen, C., Jiang, Y., Li, L.: Solo: Segmenting objects by
  locations. In: ECCV (2020)

\bibitem{wang2020solov2}
Wang, X., Zhang, R., Kong, T., Li, L., Shen, C.: Solov2: Dynamic and fast
  instance segmentation. In: NeurIPS (2020)

\bibitem{woo2018cbam}
Woo, S., Park, J., Lee, J.Y., Kweon, I.S.: Cbam: Convolutional block attention
  module. In: ECCV (2018)

\bibitem{wu2021cvt}
Wu, H., Xiao, B., Codella, N., Liu, M., Dai, X., Yuan, L., Zhang, L.: Cvt:
  Introducing convolutions to vision transformers. In: IEEE CVPR (2021)

\bibitem{xie2021segformer}
Xie, E., Wang, W., Yu, Z., Anandkumar, A., Alvarez, J.M., Luo, P.: Segformer:
  Simple and efficient design for semantic segmentation with transformers. In:
  NeurIPS (2021)

\bibitem{yan2021mirrornet}
Yan, J., Le, T.N., Nguyen, K.D., Tran, M.T., Do, T.T., Nguyen, T.V.: Mirrornet:
  Bio-inspired camouflaged object segmentation. IEEE Access  \textbf{9},
  43290--43300 (2021)

\bibitem{yang2021uncertainty}
Yang, F., Zhai, Q., Li, X., Huang, R., Luo, A., Cheng, H., Fan, D.P.:
  Uncertainty-guided transformer reasoning for camouflaged object detection.
  In: IEEE CVPR (2021)

\bibitem{zhai2021mutual}
Zhai, Q., Li, X., Yang, F., Chen, C., Cheng, H., Fan, D.P.: Mutual graph
  learning for camouflaged object detection. In: IEEE CVPR (2021)

\bibitem{zhu2021inferring}
Zhu, J., Zhang, X., Zhang, S., Liu, J.: Inferring camouflaged objects by
  texture-aware interactive guidance network. In: AAAI (2021)

\bibitem{zhu2020deformable}
Zhu, X., Su, W., Lu, L., Li, B., Wang, X., Dai, J.: Deformable detr: Deformable
  transformers for end-to-end object detection. In: ICLR (2020)

\end{thebibliography}
\end{document}